\pdfoutput=1

\documentclass[11pt]{article}

\usepackage[preprint]{acl}
\usepackage{booktabs}  
\usepackage{multirow}  
\usepackage{graphicx}  
\usepackage{adjustbox} 
\usepackage{amsmath}
\usepackage{times}
\usepackage{latexsym}
\usepackage{float}
\usepackage[T1]{fontenc}

\usepackage[utf8]{inputenc}

\usepackage{microtype}

\usepackage{inconsolata}

\usepackage{graphicx}

\title{Towards Efficient Pre-training: Exploring FP4 Precision in Large Language Models}

\author{
    Jiecheng Zhou$^{1,2}$ \quad
    Ding Tang$^2$ \quad
    Rong Fu$^2$ \footnotemark[1] \quad
    Boni Hu$^2$ \quad
    Haoran Xu$^2$ \quad
    Yi Wang$^2$ \quad
    \\
    \textbf{Zhilin Pei}$^2$ \quad
    \textbf{Zhongling Su}$^2$ \quad
    \textbf{Liang Liu}$^2$ \quad
    \textbf{Xingcheng Zhang}$^2$ \footnotemark[1] \quad
    \textbf{Weiming Zhang}$^{1}$ \footnotemark[1] \\
    $^1$School of Information Science and Technology, University of Science and Technology of China \\
    $^2$Shanghai Artificial Intelligence Laboratory
    \\
    \texttt{\{zhoujiecheng, zhangwm\}@mail.ustc.edu.cn, \{furong,zhangxingcheng\}@pjlab.org.cn}
}

\begin{document}
\maketitle

\vspace{-0.2cm}
\begin{abstract}
The burgeoning computational demands for training large language models (LLMs) necessitate efficient methods, including quantized training, which leverages low-bit arithmetic operations to reduce costs. While FP8 precision has shown potential, leveraging FP4 remains challenging due to inherent quantization errors and limited representation capability. Based on the Transformer architecture, we present an FP4 training scheme for LLMs, overcoming these obstacles through mixed-precision quantization strategies tailed for different modules and training stages. This allows us to apply the precision level suitable to distinct components within the model, ensuring that multi-head attention and linear layers are handled appropriately. Our pretraining recipe ensures stability in backpropagation by incorporating fine-grained quantization methods with a target precision training schedule. 
Experimental results demonstrate that our FP4 training scheme achieves accuracy comparable to BF16 and FP8, with smaller theoretical computational cost. With the advent of next-generation hardware supporting FP4, our method sets the foundation for efficient ultra-low precision training.
\end{abstract}

\begin{figure*}[htbp]   
  \centering
  \includegraphics[width=\textwidth]{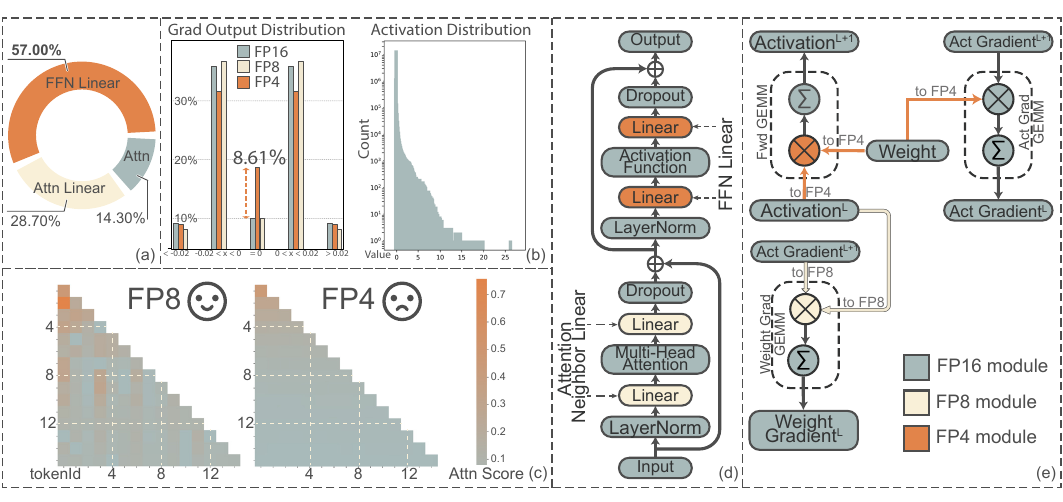}  
  \caption{(a) shows the proportion of computational overhead for the main computation components of a transformer block when using the LLaMA 7B configuration with a sequence length of 4K. (b) shows the distribution of activations and gradients after the GPT-large model has been trained to approximately 10B tokens. (c) shows the heatmap of attention scores when using different training strategies. (d) and (e) illustrate our training scheme, which will be detailed in Section \ref{sec:method}.}  
  \label{fig:teaser}  
\end{figure*}

\vspace{-0.3cm}
\section{Introduction}
\par Recent advancements in large language models, including GPT~\cite{24, 32, 33}, DeepSeek~\cite{26}, Llama~\cite{25} and OPT~\cite{35}, have demonstrated strong generalization capabilities across various tasks~\cite{36, 37}. Among these advancements, pretraining on large-scale unlabeled data has proven to be critical for ensuring model performance~\cite{24, 38}. Increasing the model size and the dataset scale can enhance performance~\cite{27, 28}, but this improvement comes with significant computational costs. To address this, numerous methods have been proposed to accelerate the pretraining process~\cite{29}.
Particularly, low-precision computation serves as an efficient acceleration technique. This approach quantizes the inputs of computationally intensive operators to a specified low-bit width, leveraging low-bit arithmetic units to speed up training.

\par Previous research on low-precision training has primarily focused on deep learning models. However, these methods do not fully consider the characteristics of large language model pre-training, which has unique training methods and model architectures. The good news is that next-generation hardware will support FP4 and FP8 format~\cite{30}. Studies like \cite{peng2023fp8, micikevicius2022fp8, 31,  fishman2024scaling, xijetfire} have demonstrated the capability of 8-bit computation for LLM pretraining. However, the application of FP4 tensor cores in LLM pretraining remains unexplored. Compared to INT4, FP4 offers a larger numerical representation space, making it possible to further reduce the bit width in large-scale model pretraining. However, the limited number of bits in FP4 format introduces significant quantization errors, making the application of FP4 to pretraining highly challenging.
\par From the perspective of LLM structure, it has been observed that different modules and computational components exhibit varying levels of sensitivity. Given the critical role of the multi-head attention (MHA) mechanism in the Transformer architecture, it is imperative to implement specific strategies to ensure the accuracy of attention modules. As the volume of data continues to grow, the gradient values tend to decrease, making FP4 quantization more prone to underflow, thus hindering parameter updates. Based on these observations, we propose a novel FP4 mixed-precision large language model pretraining recipe. Specifically, we leverage a per-block quantization strategy and employ different quantization approaches across modules and training stages to enable FP4 model training. The approaches enable better exploitation of the computational improvements brought by future hardware advancements.
\par In this paper, we explored the use of FP4 precision in language model pretraining and, for the very first time, proposed an effective mixed precious pretraining strategy.
First, considering the distinct requirements of different modules, we applied tailored quantization strategies to preserve the precision of MHA execution.
Second, as backpropagation has shown high sensitivity to precision, we employed finer-grained quantization methods to ensure accurate parameter updates during the backward pass.
Finally, we adopted a 2-stage target precision training schedule to eliminate the impact of quantization noise on the model.

\vspace{-0.2cm}
\section{Related Work}
Low-precision training enhances deep learning efficiency by reducing computational costs. Many existing studies focus on the training of deep neural networks (DNNs) \cite{wang2018training,chmiel2023accurate, sun2019hybrid,xi2023training,fu2021cpt}, whose architecture and performance differ from LLM pre-training. 
In the context of low-precision training for large model pre-training, some progress has been made in FP8. For example, \cite{micikevicius2022fp8} introduced new FP8 floating-point formats (E4M3 and E5M2), 
and \cite{fishman2024scaling} extends FP8 to trillion-token large-scale model pretraining. In terms of FP4 training, \cite{wang2025optimizing} improved FP4 computational precision using a differentiable quantization estimator and outlier clamping and compensation strategy. However, most existing methods fail to fully account for the varying sensitivity to precision across different model modules.

\section{Methods}\label{sec:method}
Our objective is to maximize the efficiency of low-precision computations based on the characteristics of LLMs. As shown in Fig.\ref{fig:teaser}(a), the computational cost of three key components in a transformer model is analyzed, with FFN accounting for 57\%. Considering both the computational cost and impact on performance, we meticulously design three corresponding training schemes: \ref{sec:attention} Attention-protected Neighbor Linear, \ref{sec:ffn} Gradient-sensitive Linear, and \ref{sec:target} Target Precious Training Schedule. These schemes fully leverage hardware acceleration while keeping precision loss within an acceptable range during training.

\vspace{-0.2cm}
\subsection{Attention-protected Neighbor Linear}\label{sec:attention}

\par As the core component of the Transformer model, the attention mechanism is highly sensitive to precision. Quantization errors introduced by low-precision training can accumulate over time, eventually disrupting the function of the attention mechanism. As shown in Fig.~\ref{fig:teaser}(c), an undisturbed attention mechanism identifies tokens 0, 3, 6, and 9 as more important. However, under FP4 training, the result becomes nearly uniformly distributed, preventing the model from distinguishing which tokens are significant. This makes it difficult for the model to differentiate the importance of tokens, thereby affecting the convergence speed.
\par To ensure the proper functioning of the attention mechanism and enable the model to correctly evaluate the importance of each token, we employ FP8 precision for the computation of QKV and the output projection to "protect" the accurate execution of the attention mechanism, as shown in Fig. \ref{fig:teaser}(d).

\vspace{-0.2cm}
\subsection{Gradient-sensitive FFN Linear}\label{sec:ffn}
\par Weight gradient computation is more sensitive to errors compared to forward computation, due to the fact that both gradients and activations contribute to the error. For gradients, since many values are around 0.02, especially as training progresses and gradient magnitudes decrease, underflow is likely to occur. As can be seen in the left of Fig.\ref{fig:teaser}(b), there is an 8.6\% difference between FP4 and FP8/FP16, thus requiring a more accurate representation. For activations, we observe that underflow occurs approximately 18\% of the time between FP4 and FP8/FP16. This is largely due to the relatively large range of values, as can be seen in the right of Fig. \ref{fig:teaser}(b). Therefore, a more accurate representation is also needed. Additionally, optimizers use the gradients to update model parameters. Based on the above discussion, for the weight gradient computation of model weights, we adopt FP8 precious computation, as shown in the bottom left corner of Fig.\ref{fig:teaser}(e).
\par Furthermore, for the activation gradient computation (the top right corner of Fig.\ref{fig:teaser}(e)), we find that quantizing gradients significantly impacts the convergence of model training. There is always a nonlinear operation between the linear layers, which requires more precise numerical representations. Furthermore, quantization errors accumulate iteratively through the chain rule during backpropagation, ultimately hindering the convergence of model training.
\par Lastly, in our experiments, we observed that quantization noise increases as the model size and the amount of data grow (a detailed explanation can be found in Appendix \ref{sec:appendixb}). This occurs because, when the model reaches a certain level of accuracy, coarse-grained low-precision tensors can no longer currently represent the parameter space and input information. Therefore, we adopt a more conservative quantization approach to maintain stable training in forward computation. As shown in the top left corner of Fig.\ref{fig:teaser}(e). To ensure efficient hardware implementation, we use per-block quantization strategies where the block size is set to 128.

\vspace{-0.2cm}
\subsection{Target Precious Training Schedule}\label{sec:target}
\begin{figure}[htbp]   
  \centering
  \includegraphics[width=0.85\linewidth]{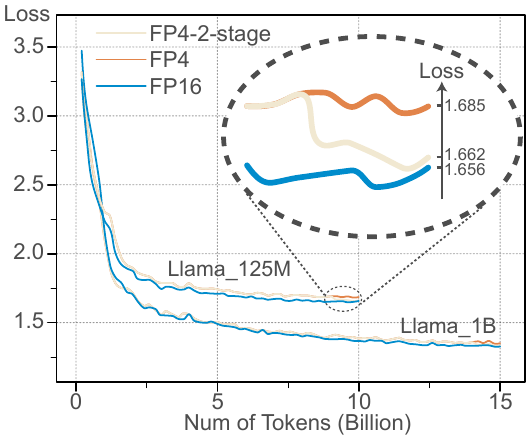}  
  \caption{Loss curve for the Target Precious Training Schedule.}  
  \label{fig:2-stage}  
\end{figure}

When using low-precision training throughout the entire process, there tends to be a performance gap between the low-precision model and the FP16 model, as shown in Fig.\ref{fig:2-stage}. The validation loss curves exhibit a parallel trend. Although the gap between the two curves is very small, the difference in downstream tasks, such as wikitext perplexity (PPL), can be more pronounced, reaching up to approximately 6.3 compared to the model trained with FP16. This is likely due to compromises the model makes to adapt to the noise introduced by quantization during low-precision training. 
To address this issue, we employ a Target Precious Training Schedule, which involves two stages: continuing the FP4 pretraining process with FP16 for a short period. This accounts for only 5\% to 10\% of the total training steps, allowing the model to return to an ideal state. 
\begin{table*}[h!]
    \centering
    \renewcommand{\arraystretch}{1.4} 
    \setlength{\tabcolsep}{4pt} 
    \caption{Comparison of FP4 and FP16 Training Results}
    \label{tab:results}
    \resizebox{\textwidth}{!}{
    \begin{tabular}{llc|c|c|ccccccccc}
        \toprule
        Model & Method & Val Loss & Val PPL & \multicolumn{2}{l}{Text Gen} & \multicolumn{7}{c}{Natural Language Understanding (GLUE)} \\
        & & & & WikiText & cola & sst2 & mrpc & stsb & rte & wnli & qnli & mnli & qqp \\
        \midrule
        \multirow{2}{*}{GPT2 125M} & Ours & 1.706 & 5.507 & 50.98 & 0.2663 & 0.8704 & 0.7549 / 0.8322 / 0.7936 & 0.7681 / 0.7658 & 0.5704 & 0.3380 & 0.8473 & 0.7554 / 0.7705 & 0.8777 / 0.8403 \\
        & FP16-baseline & 1.705 & 5.503 & 50.14 & 0.2290 & 0.8796 & 0.7647 / 0.8395 / 0.8021 & 0.7798 / 0.7808 & 0.5884 & 0.3099 & 0.8548 & 0.7613 / 0.7695 & 0.8799 / 0.8437 \\
        \midrule
        \multirow{2}{*}{GPT2 335M} & Ours & 1.549 & 4.705 & 37.62 & 0.2565 & 0.8899 & 0.7647 / 0.8362 / 0.8004 & 0.8159 / 0.8122 & 0.6029 & 0.2535 & 0.8611 & 0.7798 / 0.7874 & 0.892 / 0.8572 \\
        & FP16-baseline & 1.556 & 4.739 & 38.39 & 0.3002 & 0.8819 & 0.7745 / 0.8419 / 0.8082 & 0.8266 / 0.8298 & 0.6209 & 0.1831 & 0.8726 & 0.7799 / 0.7889 & 0.8929 / 0.8508 \\
        \midrule
        \multirow{2}{*}{GPT2 774M} & Ours & 1.431 & 4.181 & 30.01 & 0.3473 & 0.9002 & 0.7745 / 0.8472 / 0.8108 & 0.8305 / 0.8336 & 0.6498 & 0.2676 & 0.8742 & 0.7995 / 0.808 & 0.8984 / 0.8656 \\
        & FP16-baseline & 1.430 & 4.178 & 28.36 & 0.3708 & 0.8922 & 0.7794 / 0.8454 / 0.8124 & 0.8347 / 0.8353 & 0.6498 & 0.2254 & 0.8911 & 0.8078 / 0.813 & 0.9012 / 0.8683 \\
        \bottomrule
    \end{tabular}
    }
\end{table*}

\vspace{-0.5cm}
\section{Experiment}
In this section, we evaluate the proposed FP4 training method across language models of various sizes. The detailed model training configurations and hyperparameter settings are provided in Appendix \ref{sec:appendixb}. Section 4.1 presents the main results, showcasing the model's performance on downstream tasks. Section 4.2 presents the ablation study to demonstrate the effectiveness of our training method.
\vspace{-0.2cm}
\subsection{Main Result}\label{main_results}
We validate the proposed FP4 pretraining method on two large language models, using the widely adopted GPT-2 and LLaMA architectures. The GPT-2 and LLaMA models are pretrained on the RedPajama-WikiText~\cite{weber2025redpajama} dataset within the Megatron framework and evaluate their text generation capabilities on wikiText\cite{merity2016pointer}. Additionally, we assess their natural language understanding abilities on the GLUE\cite{wang2018glue} benchmark.

We train approximately 10B tokens on GPT-2-small and GPT-2-mid and around 25B tokens on GPT-2-large. The final validation loss and validation perplexity (PPL) are presented in Table \ref{tab:results}, showing that the pretraining results obtained with our method exhibit almost no performance difference compared to models trained using FP16.
In addition to training loss, the downstream task performance of the same pretrained models demonstrates that the average accuracy of FP4-trained models is comparable to that of FP16-trained models.
\subsection{Ablation Study}
We aim to investigate the effect of the module-wise pretraining method introduced in Section \ref{sec:method}. For this ablation study, we train the LLaMA2-125M model on approximately 5B tokens. The results in the table indicate that different modules exhibit varying levels of robustness to low precision. Additionally, we compute the theoretical computation cost for these methods and observe that our approach achieves a lower theoretical computation cost (see Appendix \ref{sec:appendixb} for details) while maintaining higher performance.
\begin{table}[H]
    \resizebox{\columnwidth}{!}{%
    \Large
    \setlength{\tabcolsep}{4pt} 
    \renewcommand{\arraystretch}{1.2} 
    \begin{tabular}{c|c|c|c|c|c|c}
        \toprule
        {Attention Linear} & {FFN Linear} & {FP4 Linear' Backward} & {Training loss} & {Val loss} & {Val ppl} & {Computation cost} \\[4mm]
        \midrule
        FP4  & FP4  & FP4  & 2.2659 & 1.7828 & 5.9467  & 57.1\%  \\
        FP4  & FP8  & FP8  & 2.2211 & 1.7543 & 5.7798  & 69.6\%  \\
        FP8  & FP4  & FP4  & 2.2562 & 1.7549 & 5.7831  & 60.7\%  \\
        \textbf{FP8}  & \textbf{FP4}  & \textbf{FP8}  & \textbf{2.2225} & \textbf{1.7415} & \textbf{5.7062}  & \textbf{66.1}\%  \\
        FP16 & FP16 & FP16 & 2.1998 & 1.7097 & \textbf{5.5273}  & 100\%   \\
        \bottomrule
    \end{tabular}%
    }
    \caption{Ablation studies about different precious on different modules}
    \label{tab:result}
\end{table}
\begin{table}[H]
    \resizebox{\columnwidth}{!}{%
    \centering
    \setlength{\tabcolsep}{4pt} 
    \renewcommand{\arraystretch}{1.2} 
    \begin{tabular}{lccccccc}
        \toprule
        \multirow{2}{*}{} & \multirow{2}{*}{{Attention}} & \multirow{2}{*}{{FFN}} & \multirow{2}{*}{{FFN Backward}} & \multirow{2}{*}{{Target Precious}} & \multirow{2}{*}{{Val loss}} & \multirow{2}{*}{{Val ppl}} & \multirow{2}{*}{{Computation}} \\
        & & & & & & & {Cost} \\
        \midrule
        \multirow{3}{*}{\textbf{Llama 1B}}  
        & FP8  & FP4  & FP8  & no  & 1.3505 & 3.8596  & 67.5\%  \\
        & FP8  & FP4  & FP8  & yes & 1.3311 & \textbf{3.7855}  & 69.7\%  \\
        & FP16 & FP16 & FP16 & -   & 1.3296 & \textbf{3.7797}  & 100\%   \\
        \midrule
        \multirow{3}{*}{\textbf{Llama 125M}}  
        & FP8  & FP4  & FP8  & no  & 1.6851 & 5.3933  & 68.2\%  \\
        & FP8  & FP4  & FP8  & yes & 1.6622 & \textbf{5.2670}  & 71.4\%  \\
        & FP16 & FP16 & FP16 & -   & 1.6567 & \textbf{5.2424}  & 100\%   \\
        \bottomrule
    \end{tabular}%
    }
    \caption{Ablation studies about target precious training schedule}
    \label{tab:llama_results}
\end{table}
\vspace{-0.3cm}
Furthermore, to demonstrate the effectiveness of the 2-stage training schedule, we conducted the following ablation experiment, using the same experimental setup as in Section \ref{main_results}. The results in the table highlight the importance of the 2-stage training schedule for large-scale model pretraining.

\vspace{-0.2cm}
\section{Conclusion}
We propose an FP4 pre-training scheme for modern large language models. 
Based on the sensitivity analysis of computational modules and training stages, we adopt a tailored training recipe according to the position of linear modules, together with a Target Precious Training schedule, to ensure stable convergence. 
Experimental results show that FP4-based training achieves comparable validation loss and downstream task accuracy to traditional FP16 training while reducing computational costs by 30\%. Additionally, our ablation studies confirm the importance of adaptive quantization strategies across different model modules and training stages, providing new insights for the further development of low-precision training techniques and efficient training of large language models on next-generation hardware.
\section{Limitation}
First, due to computational resource limitations, our method has not been validated on larger models and larger datasets to demonstrate its effectiveness. Investigating such scalability remains a critical direction for future research. In addition, since the model adopts a simulated FP4 approach, it is unable to obtain an accurate increase in training efficiency. Lastly, for the sensitive computational components, we employed a simple strategy to ensure numerical precision. In future work, we will explore more customized approaches to enable a broader range of computations to utilize FP4.

\bibliography{custom}
\clearpage
\appendix

\section{FP4 Quantization}\label{sec:appendixa}
Quantization is the process of converting a data type with more bits (e.g., 32- or 16-bit floating points) into another data type with fewer bits (e.g., 4-bit floating points). In integer quantization, the real-valued variable $X_R$ is quantized to an integer $X_{INT}$ with the following formula:
\begin{equation}
X_{\text{INT}} = \alpha \left\lfloor \text{Clip}\left(\frac{X_R}{\alpha}, Q_{\text{min}}, Q_{\text{max}}\right) \right\rceil
\end{equation}

\par Similar to integer quantization, in float point quantization, scaling and clipping of the values are required before quantization, as follows. 
\begin{equation}
Q_{\text{max}} = -Q_{\text{min}} = (2 - 2^{-m}) 2^{2^e - b - 1}
\end{equation}
\begin{equation}
\tilde{Q}_{\text{max}} = \alpha Q_{\text{max}}
\end{equation}
\begin{equation}
X_R' = \text{Clip}\left(X_R, \tilde{Q}_{\text{min}}, \tilde{Q}_{\text{max}}\right)
\end{equation}
\par Where the min/max value range of signed floating-point quantization can be calculated from Eq. (2), and the scaling factor $\alpha$ determines the quantization granularity. Thus, we can determine the upper and lower bounds of floating-point quantization and perform the clip operation accordingly.
\par After scaling and clipping, we can quantize the real value into a specific data format. Unlike INT quantization, floating-point numbers have different quantization levels for different values. Therefore, we first need to determine the quantization step size:
\begin{equation}
\tilde{\alpha} = 2^{-\tilde{b}} = 2^{-b} \cdot \alpha
\end{equation}
\begin{equation}
v = 
\begin{cases} 
    2^{\lfloor \log_2 |X_R'| /\tilde{\alpha}  \rceil - m} & \text{if } \lfloor \log_2 |X_R'| /\tilde{\alpha}  \rceil \geq 1 \\
    2^{1-m} & \text{otherwise}
\end{cases}
\end{equation}
Here, the quantization level v is determined by $x_R' / \tilde{\alpha}$, after which the floating-point number can be quantized following the format of Eq. (8). A detailed explanation can be found in \cite{13, 12}. Finally, the quantization formula can be expressed as follows:
\begin{equation}
    X_{\text{FP}} = \tilde{\alpha} \cdot v \cdot \left\lfloor \frac{X'_R}{\tilde{\alpha} \cdot v} \right \rceil
\end{equation}

\section{Training Detail}\label{sec:appendixb}
We conducted our experiments on the Megatron~\cite{19} framework and evaluated downstream performance using the transformers~\cite{wolf-etal-2020-transformers} and lm-evaluation-harness~\cite{eval-harness}, ensuring standardized and reproducible benchmarking. Hyperparameters remain consistent across precision settings for fair comparison. The learning rate follows a warm-up and cosine decay schedule, with the warm-up phase spanning 0.15\% of total steps and the learning rate gradually decreasing to 10\% of its peak over the remaining 90\%. The peak learning rate is $1 \times 10^{-4}$, with a weight decay of 0.1 for Llama models. For GPT models, the peak learning rate is set to $6 \times 10^{-4}$, with a weight decay of 0.1 For the Adam optimizer, we use $\beta_1 = 0.9$, $\beta_2 = 0.95$, and $\epsilon = 1 \times 10^{-8}$. For llama models, the input sequences are fixed at 2048 tokens, and the batch size is 512, comprising approximately 1M tokens. For GPT models, the input sequences are fixed at 1024 tokens, and the batch size is 480, comprising approximately 0.5M tokens, detail model config can be found in Table \ref{tab:model_config}.
\begin{table}[h]
    \resizebox{\columnwidth}{!}{%
    \centering
    \renewcommand{\arraystretch}{1.2} 
    \setlength{\tabcolsep}{8pt} 
    \begin{tabular}{lccccc}
        \toprule
        \textbf{Parameter} & \textbf{GPT-125M} & \textbf{GPT-335M} & \textbf{GPT-774M} & \textbf{LLaMA-125M} & \textbf{LLaMA-1B} \\
        \midrule
        Layers                & 12  & 24  & 36  & 12  & 48  \\
        Hidden Size           & 768 & 1024 & 1280 & 768 & 1280 \\
        Activation Function   & GELU & GELU & GELU & SwiGLU & SwiGLU \\
        Normalization         & LayerNorm & LayerNorm & LayerNorm & RMSNorm & RMSNorm \\
        FFN Hidden Size       & 3072 & 4096 & 5120 & 3072 & 3392 \\
        Sequence Length       & 1024 & 1024 & 1024 & 2048 & 2048 \\
        Attention Heads       & 12  & 16  & 20  & 12  & 20  \\
        \bottomrule
    \end{tabular}%
    }
    \caption{GPT and LLaMA Model Configurations}
    \label{tab:model_config}
\end{table}
\par We quantize all the linear layers in the MLP and attention
module to target precious, and leave multi-head attention and activation function in FP16 by employing FlashAttention~\cite{20}. The master copy of the weights is kept in FP32. We quantize
linear layers’ inputs to target precious prior to each matmul, but
leave layernorm’s weight and bias to floating-point since
they are relatively small.
\par For the calculation of theoretical computation cost, we first separately count the forward and backward computation amounts for each part of a Transformer block (considering only computations related to matrix multiplications, as these account for more than 95\% of the total computation). Then, based on the assumptions that FP8 achieves twice the computation speed of FP16 and FP4 achieves four times the computation speed of FP16, we compute the theoretical time required for each matrix multiplication. Finally, we obtain the theoretical computation cost for each method.
\par Since the quantized weight $\tilde{w}$ is an estimate of $w$, 
We directly use a straight-through estimator~\cite{17} directly passes the gradient of $\tilde{w}$ to $w$: \[ \nabla_{\mathbf{w}} \mathcal{L}(\tilde{\mathbf{w}}) \gets \nabla_{\tilde{\mathbf{w}}} \mathcal{L}(\tilde{\mathbf{w}}). \]
\par In our experiments, we found that different models have varying precision requirements. For the GPT-125M model, applying a per-token and per-channel FP4 quantization strategy for both forward computation and weight gradient computation is feasible. The final results are shown in Table \ref{tab:results}. The use of per-token and per-channel quantization is designed to better align with matrix multiplication rules, allowing for efficient implementation on accelerators. However, for the GPT-335M model, the per-token and per-channel FP4 quantization strategy becomes ineffective as the data volume increases. In this case, switching to per-block FP4 quantization for weight gradient computation enables training to proceed, with the final results also presented in Table \ref{tab:results}. For the GPT-770M model, the quantization strategy used for GPT-335M becomes ineffective as training progresses. At this point, modifying the forward computation to use per-block FP4 quantization while increasing the precision of weight gradient computation to FP8 ensures stable training. Additionally, we validated the feasibility of the GPT-770M quantization strategy (as we discuss in Section \ref{sec:method}) on the LLaMA-125M and LLaMA-1B models. It can be anticipated that as model size and data volume continue to grow, the precision requirements for model training will become increasingly stringent\cite{11}. Ensuring that FP4 can support long-term, large-scale pretraining of large models remains a key direction for our future work.
\end{document}